\documentclass{article}
    
    \usepackage{amsmath,amssymb}
    \usepackage{graphicx}
    \usepackage{booktabs}
    \usepackage{cite}
    \usepackage{mlspconf}
    \usepackage{microtype}
    \usepackage{xurl}
    \usepackage[hidelinks]{hyperref}    
    \copyrightnotice{979-8-3195-0884-3/26/\$31.00 {\copyright}2026 IEEE}
    
    \toappear{2026 IEEE International Workshop on Machine Learning for Signal Processing,
    Sep.\ 28--Oct.\ 1, 2026, Atlanta, USA}
    
    \title{MSR-IVA: Masked Structural Residual Independent Vector Analysis for State-Aware Fusion of Structural MRI and Dynamic Functional Network Connectivity}
    \name{
    Victor Solomon$^{*\dagger}$ \qquad
    Zening Fu$^{*\dagger}$ \qquad
    Rafal Angryk$^{*\dagger}$ \qquad
    Vince D. Calhoun$^{*\dagger}$ \qquad
    Jingyu Liu$^{*\dagger}$
    }
    
    \address{
    $^{*}$Department of Computer Science, Georgia State University\\
    $^{\dagger}$Tri-Institutional Center for Translational Research in Neuroimaging and Data Science (TReNDS),\\
    Georgia State University, Georgia Institute of Technology, and Emory University
    }
    
\begin{document}
    \maketitle

    
    \begin{abstract}
    Multimodal fusion of structural MRI (sMRI) and dynamic functional network connectivity (dFNC) can reveal how brain structure relates to changing functional states. When the same structural latent representation is coupled with multiple states, applying independent vector analysis (IVA) separately to each state can produce unrelated structural decompositions, while forcing identical decompositions may suppress state-specific relationships. In addition, not every subject expresses every dynamic state. We propose masked structural residual IVA (MSR-IVA), a state-aware framework that combines a shared structural representation with state-specific residual adaptations and masks for incomplete state expression. On an Alzheimer's Disease Neuroimaging Initiative cohort, MSR-IVA improved matched source coupling by 6.5\% and reduced unmatched dependence by 15.7\% relative to the independent pairwise IVA baseline. Among subjects expressing both states, mean absolute cross-state structural source correlation was \(0.9177\) for MSR-IVA versus \(0.2978\) for no sharing, demonstrating controlled structural sharing that preserves source correspondence while allowing state-specific adaptation.
    \end{abstract}
    
    
    \section{Introduction}
    
    Multimodal neuroimaging combines complementary information from different imaging modalities to provide a more comprehensive understanding of the brain. Structural MRI (sMRI) captures gray matter morphology, while resting-state fMRI (rs-fMRI) measures functional interactions among distributed brain networks. The value of preserving and analyzing temporal structure has also been
    demonstrated in other multivariate time series settings
    \cite{solomon2025timeseries}. Dynamic functional network connectivity
    (dFNC) further characterizes recurring connectivity states that reflect
    changes in functional organization over time \cite{allen2014tracking}. Integrating structural and dynamic functional information can therefore help characterize how anatomical variation in the brain relates to different functional connectivity states. This is particularly relevant in Alzheimer's disease (AD), where alterations in both brain structure and functional network organization have been reported \cite{sendi2021alzheimer}.
    
    Source-based multimodal fusion methods such as joint ICA and independent vector analysis (IVA) can identify linked latent sources across multiple datasets \cite{adali2015multimodal}. IVA organizes corresponding sources into source component vectors (SCVs), allowing statistical dependence among matched sources while separating distinct SCVs. This property makes IVA suitable for identifying linked structural and functional representations.
    
    Existing fusion approaches do not directly address scenarios in which a structural representation is coupled with multiple state-specific dFNC representations. A direct approach is to fit independent pairwise IVA models between sMRI and each dFNC state, following a conventional IVA formulation similar to \cite{silva2021misa}. While this allows each state to be modeled separately, it produces independent structural decompositions and does not preserve correspondence across states. In contrast, enforcing a single identical structural decomposition across all states may be overly restrictive because structural relationships with functional connectivity can vary across dynamic states.
    
    A second challenge arises from incomplete state expression. A subject may express one dynamic state but not another during the acquired scan. Requiring complete observations across all states would therefore discard subjects who still contain valid information for individual state-specific fusion problems. A state-aware formulation should use all available observations while preventing absent states from contributing to the corresponding objective.
    
    We propose masked structural residual independent vector analysis (MSR-IVA), a state-aware IVA framework for fusing one structural modality with multiple partially observed dynamic functional states. MSR-IVA couples state-specific structural representations through a shared component with residual adaptations, while state masks ensure that each sMRI--dFNC fusion problem uses only subjects with valid observations for that state. This provides a soft sharing formulation between fully independent and fully shared structural decompositions.
    
    Our contributions are threefold. First, we introduce a masked IVA formulation that accommodates incomplete state expression without requiring complete cases. Second, we develop a structural residual parameterization that preserves cross-state correspondence while allowing state-specific adaptation. Third, we evaluate MSR-IVA using source coupling, cross-SCV dependence, structural source correspondence, residual regularization, repeated random initializations, and SCV label permutation analysis.

    
    \section{Data and Preprocessing}
    
    \subsection{Data and Cohort Construction}
    \label{subsec:data_cohort}
    
    Data were obtained from the Alzheimer's Disease Neuroimaging Initiative (ADNI) \cite{jack2008alzheimer}. The rs-fMRI and T1-weighted sMRI were matched at the subject level. We used the ADNI rs-fMRI subset acquired with a repetition time of 3 seconds. Diagnosis groups included cognitively normal (CN), mild cognitive impairment (MCI), and dementia.
    
    For each fMRI scan, the closest clinical visit within 180 days was selected, followed by the closest available T1-weighted gray matter image within 180 days of that visit. The matched cohort contained 573 subjects, including 296 CN, 160 MCI, and 117 dementia subjects. State~1 was expressed by 363 subjects, including 192 CN, 100 MCI, and 71 dementia subjects, while State~2 was expressed by 219 subjects, including 98 CN, 69 MCI, and 52 dementia subjects. Only 124 subjects expressed both selected states.
    
    Overall, 458 subjects expressed at least one of States~1 or 2, while the remaining 115 did not contribute to the present two-state fusion objective. A complete case analysis would retain only the 124 common subjects, discarding 239 State~1 subjects (65.84\%) and 95 State~2 subjects (43.38\%). The masked formulation instead retains all subjects with valid observations within each state-specific objective. The structural representation contained 66 sMRI features, while each state-specific dFNC representation contained 1378 connectivity features derived from 88 windows per scan.
    
    \subsection{Dynamic Functional Connectivity Features}
    \label{subsec:dfnc_features}
    
\begin{figure}[t]
        \centering
        \includegraphics[width=\columnwidth]{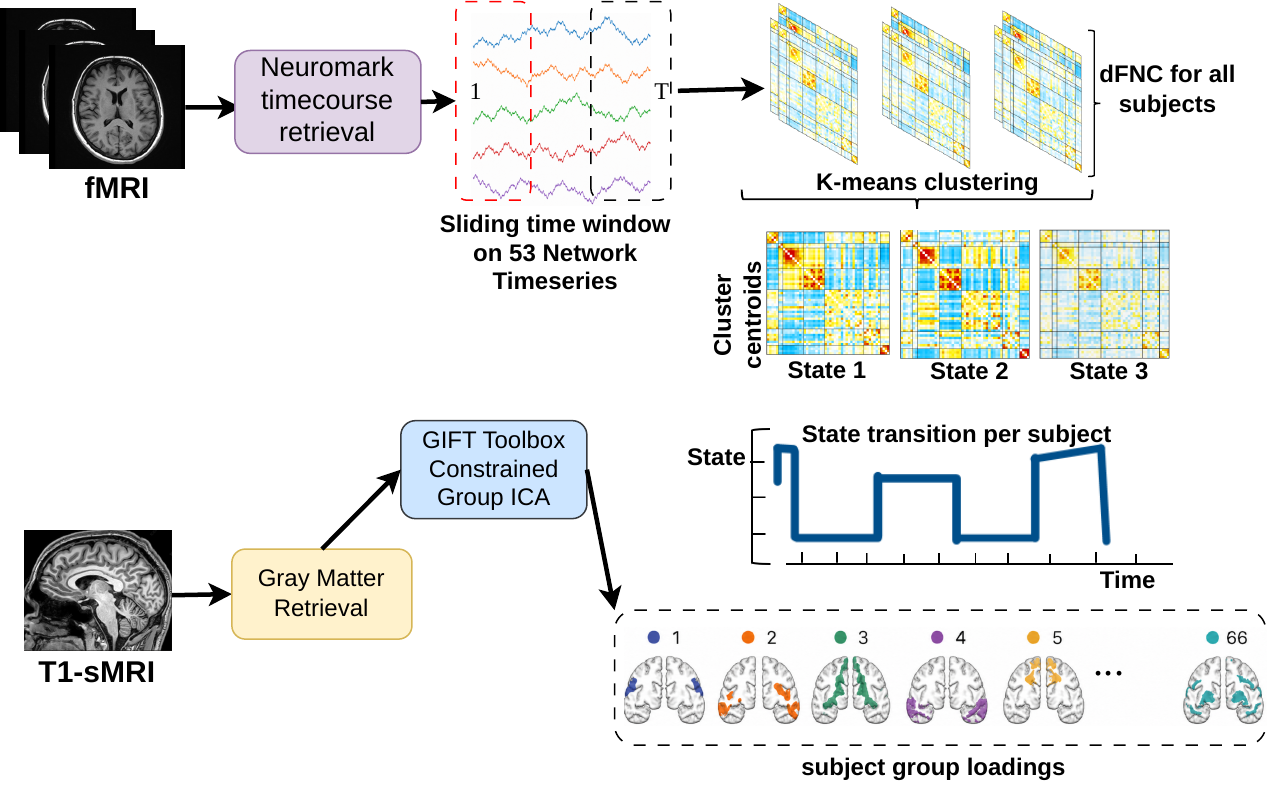}
        \caption{Overview of structural and dynamic functional connectivity processing.}
        \label{data_processing_workflow}
    \end{figure}

    The data processing workflow is summarized in Fig.~\ref{data_processing_workflow}. The rs-fMRI was processed using the NeuroMark functional pipeline \cite{du2020neuromark}, in which template-guided ICA estimated 53 subject-specific functional networks and their time courses. Sliding window dFNC was computed from pairwise connectivity among these networks using a 21-time-point window, producing 88 windows per scan and 1378 unique connectivity features per window.
    
    A group-level three-state solution was obtained using \(k\)-means clustering of the windowed dFNC vectors. The present study focused on States~1 and 2, which were expressed by 363 and 219 subjects, respectively, with only 124 subjects expressing both states. State~3 was expressed by 544 subjects and was not included because its near-complete availability provided limited opportunity to evaluate the masking strategy under missing state expression. For each subject, windows assigned to a given state were averaged to obtain a subject-level dFNC representation for that state. Subjects who did not express a given state did not contribute to the corresponding state-specific objective.
    
    \subsection{Structural MRI Features}
    \label{subsec:smri_features}
    
    The sMRI data were represented using NeuroMark source-based morphometry features. T1-weighted images were processed using a voxel-based morphometry NeuroMark pipeline \cite{du2018neuromark}. Gray matter maps were projected onto the NeuroMark 3.0 structural MRI template using constrained group ICA with MOO-ICAR \cite{du2020neuromark}, producing 66 subject-specific gray matter loadings. These loadings formed the structural feature matrix used by all fusion models.

    
    \section{State-Aware IVA Framework}
    \label{sec:state_aware_iva}
    
    IVA is a joint blind source separation method that extends independent component analysis from a single dataset to multiple related datasets \cite{kim2006iva}. Let \(\mathbf Z_k\) denote the \(k\)th SCV and \(K\) the total number of SCVs. Sources belonging to the same SCV are allowed to be statistically dependent, whereas distinct SCVs are optimized to be statistically independent. IVA can therefore be expressed as minimizing the mutual information among SCVs,
    \(I(\mathbf Z_1,\ldots,\mathbf Z_K)=\sum_{k=1}^{K}h(\mathbf Z_k)-h(\mathbf Z_1,\ldots,\mathbf Z_K)\),
    where \(h(\cdot)\) denotes differential entropy. This dependence structure is useful for multimodal fusion because corresponding latent sources can remain coupled across datasets \cite{adali2015multimodal}.

    \subsection{State-Aware sMRI--dFNC Fusion Problem}
    \label{subsec:state_fusion_problem}
    
    Let \(\mathbf X^{g}\in\mathbb R^{D_g\times N}\) denote the structural feature matrix and \(\mathbf X^{f_r}\in\mathbb R^{D_{f_r}\times N}\) the dFNC feature matrix for state \(r\in\{1,2\}\). To represent incomplete state expression, each state has a binary mask \(\mathbf m^{(r)}\in\{0,1\}^{N}\), with \(m_n^{(r)}=1\) when subject \(n\) has a valid observation for state \(r\). The corresponding valid subject set is \(\Psi_r=\{n:m_n^{(r)}=1\}\).
    
    For state \(r\), the model estimates a structural demixing matrix \(\mathbf W^{g_r}\in\mathbb R^{K\times D_g}\) and a dFNC demixing matrix \(\mathbf W^{f_r}\in\mathbb R^{K\times D_{f_r}}\). The source matrices are
    \(\mathbf Y^{g_r}=\mathbf W^{g_r}\mathbf X^{g}\) and
    \(\mathbf Y^{f_r}=\mathbf W^{f_r}\mathbf X^{f_r}\).
    Restricting the \(k\)th rows to \(\Psi_r\) gives
    \(\tilde{\mathbf y}^{g_r}_k=\mathbf y^{g_r}_{k,\Psi_r}\) and
    \(\tilde{\mathbf y}^{f_r}_k=\mathbf y^{f_r}_{k,\Psi_r}\).
    The state-specific SCV is
    \(\mathbf Z_k^{(r)}=(\tilde{\mathbf y}^{g_r}_k,\tilde{\mathbf y}^{f_r}_k)^\top\).

    \subsection{Independent Pairwise IVA Baseline}
    \label{subsec:ipiva}
    
    Independent pairwise IVA (IP-IVA) serves as an established multimodal fusion baseline and follows the conventional IVA formulation represented within the MISA framework \cite{silva2021misa}. IP-IVA fuses sMRI separately with each dFNC state. One model is estimated for \((\mathbf X^g,\mathbf X^{f_1})\) and another independently for \((\mathbf X^g,\mathbf X^{f_2})\). Each state therefore receives its own structural and functional demixing matrices. This allows each state to have its own structural decomposition but does not explicitly connect the representations learned by the two fusion problems. IP-IVA uses a multivariate Kotz source prior. For an SCV sample \(\mathbf z\), let \(\mathbf D\) denote the stabilized scale matrix used by the source model and define
    \(\delta=\mathbf z^\top\mathbf D^{-1}\mathbf z\).
    For numerical stability, \(\tilde{\delta}=\max(\delta,\epsilon_{\mathrm K})\). Up to constants, the implemented negative log density is
    
    \begin{equation}
    \ell_{\mathrm K}(\mathbf z)
    =
    \lambda\tilde{\delta}^{\beta}
    -
    (\eta-1)\log\tilde{\delta}
    +
    \frac{1}{2}\log|\mathbf D|.
    \label{eq:kotz}
    \end{equation}
    
    Because the demixing matrices are rectangular, the conventional square matrix log determinant is replaced by the log volume
    \(\mathcal V(\mathbf W)=\sum_{i=1}^{K}
    \log[\max(\sigma_i(\mathbf W),\epsilon_{\mathrm V})]\),
    where \(\sigma_i(\mathbf W)\) is the \(i\)th singular value. The IP-IVA objective for state \(r\) is
    
    \begin{equation}
    \begin{aligned}
    \mathcal L_{\mathrm{IP}}^{(r)}
    &=
    \frac{1}{K}
    \sum_{k=1}^{K}
    \ell_{\mathrm K}\!\left(\mathbf Z_k^{(r)}\right)\\
    &\quad
    -\frac{\rho_{\mathrm V}}{K}
    \left[
    \mathcal V(\mathbf W^{g_r})
    +
    \mathcal V(\mathbf W^{f_r})
    \right].
    \end{aligned}
    \label{eq:ipiva_loss}
    \end{equation}
    
    The two state-specific IP-IVA objectives are optimized independently. Consequently, although the same structural modality is supplied to both models, the structural decompositions are independently estimated.

    \subsection{Proposed Masked Structural-Residual IVA}
    \label{subsec:msriva}
    
    MSR-IVA replaces the independently estimated structural transformations with a shared plus residual parameterization,
    \(\mathbf W^{g_r}=\mathbf W^{g_0}+\Delta\mathbf W^{g_r}\),
    where \(\mathbf W^{g_0}\in\mathbb R^{K\times D_g}\) is shared across states and \(\Delta\mathbf W^{g_r}\in\mathbb R^{K\times D_g}\) is the residual adjustment for state \(r\). Each dFNC state retains its own functional transformation \(\mathbf W^{f_r}\).
    
    The resulting objective is
    
    {\setlength{\abovedisplayskip}{4pt}
    \setlength{\belowdisplayskip}{4pt}
    \begin{equation}
    \begin{aligned}
    \mathcal L_{\mathrm{MSR}}
    &=
    \frac{1}{2}
    \sum_{r=1}^{2}
    \bigg[
    \frac{1}{K}\sum_{k=1}^{K}
    \ell_{\mathrm K}\!\left(\mathbf Z_k^{(r)}\right)\\
    &\quad
    -\frac{1}{K}
    \left\{
    \rho_g\mathcal V\!\left(
    \mathbf W^{g_0}+\Delta\mathbf W^{g_r}
    \right)
    +
    \rho_f\mathcal V(\mathbf W^{f_r})
    \right\}
    \bigg]\\
    &\quad
    +
    \frac{\alpha}{2K}
    \sum_{r=1}^{2}
    \left\|
    \Delta\mathbf W^{g_r}
    \right\|_F^2,
    \qquad \alpha>0 .
    \end{aligned}
    \label{eq:msr_objective}
    \end{equation}
    }
    
    Each \(\mathbf Z_k^{(r)}\) contains only subjects in \(\Psi_r\), so absent states do not contribute to their corresponding objective. The Kotz terms model dependence within matched SCVs, the log volume terms discourage rank collapse, and the residual penalty regulates the magnitude of state-specificstructural adaptation.
    
    The parameter \(\alpha\) controls the strength of structural sharing. Smaller positive values permit larger state-specificresiduals, whereas increasing \(\alpha\) increasingly constrains the state-conditioned structural decompositions toward the shared component. We additionally implement two limiting references. The no sharing model uses independent structural matrices for the two states, whereas the hard sharing model (HSMSR-IVA) removes the state-specificstructural residuals so that both states use the same structural matrix.
    
    \begin{figure*}[t]
    \centering
    \includegraphics[width=\textwidth]{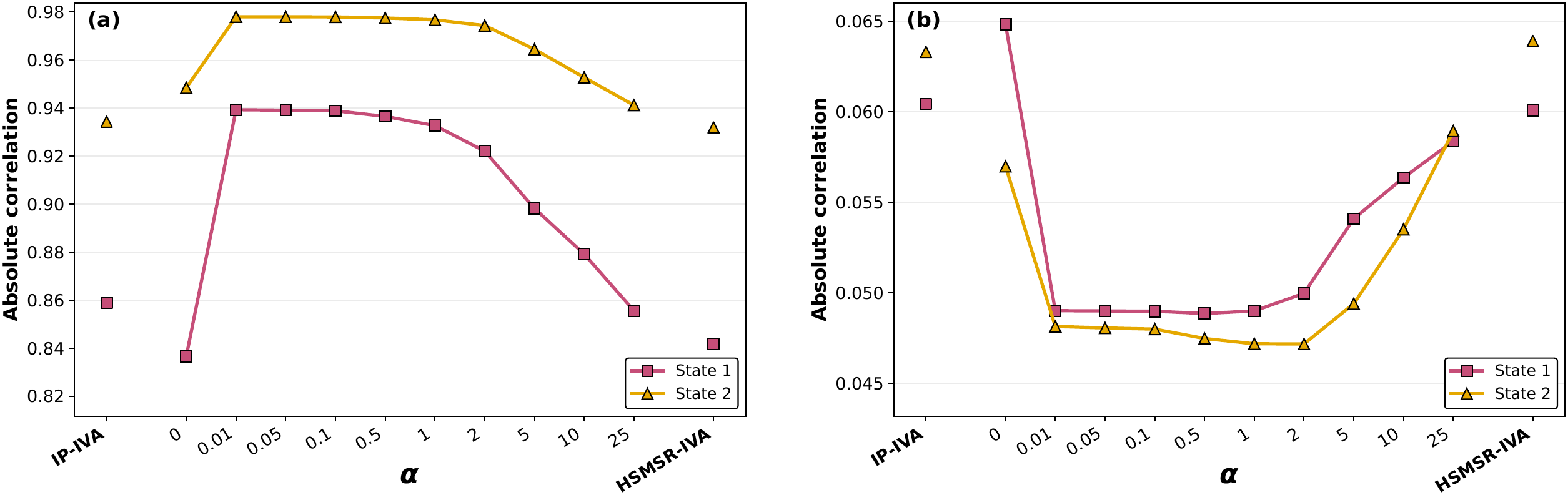}
    \caption{Sensitivity of MSR-IVA to the structural residual weight \(\alpha\). (a) Within-SCV coupling across dynamic states, where higher values indicate stronger matched source coupling. (b) Cross-SCV dependence, where lower values indicate weaker dependence between unmatched sources.}
    \label{fig:s2_sensitivity_ablation}
    \end{figure*}

    \subsection{Evaluation Metrics}
    \label{subsec:evaluation_metrics}
    
    Model quality was evaluated using absolute Pearson correlations between structural and dFNC sources over the valid subject set \(\Psi_r\). Let \(a_{kj}^{(r)}=|\rho_{kj}^{(r)}|\), where \(\rho_{kj}^{(r)}\) is the correlation between structural source \(k\) and dFNC source \(j\). Matched source coupling is \(C_{\mathrm{within}}^{(r)}=K^{-1}\sum_k a_{kk}^{(r)}\), cross-SCV dependence is \(C_{\mathrm{cross}}^{(r)}=[K(K-1)]^{-1}\sum_{k\ne j}a_{kj}^{(r)}\), and coupling contrast is \(C_{\mathrm{contrast}}^{(r)}=C_{\mathrm{within}}^{(r)}-C_{\mathrm{cross}}^{(r)}\). Higher within-SCV coupling and coupling contrast and lower cross-SCV dependence indicate better source separation and correspondence.
    
    To directly evaluate how strongly the structural representations were shared across states, we used only the common subject set \(\Psi_{12}=\Psi_1\cap\Psi_2\), containing 124 subjects. Cross-state structural similarity was \(C_{\mathrm{struct}}=\max_{\pi}K^{-1}\sum_k|\operatorname{corr}(\mathbf y^{g_1}_{k,\Psi_{12}},\mathbf y^{g_2}_{\pi(k),\Psi_{12}})|\), where \(\pi\) is a one-to-one assignment maximizing cross-state source similarity. Unlike the source coupling metrics, \(C_{\mathrm{struct}}\) is used to characterize the degree of structural sharing rather than as a higher-is-better performance measure.
    
    We additionally performed an SCV label permutation test with \(B=10{,}000\) random derangements of the dFNC SCV labels while keeping subjects fixed. SCV-wise empirical \(p\)-values were \(p_k=[1+\sum_b\mathbb I(a_{k,\pi_b(k)}^{(r)}\ge a_{kk}^{(r)})]/(B+1)\). Family-wise error was controlled using the maximum mismatched SCV correlation from each permutation.

    \subsection{Experimental Setup}
    \label{subsec:experimental_setup}
    
    \subsubsection{Baselines and Ablations}
    
    We compared MSR-IVA with IP-IVA, an established multimodal fusion baseline following the conventional IVA formulation represented within MISA \cite{silva2021misa}, the separately implemented no sharing model, and HSMSR-IVA. IP-IVA uses the same Kotz SCV dependence model and rectangular log volume formulation as MSR-IVA but estimates each sMRI--dFNC pair independently. The no sharing model provides the unconstrained structural reference within the state-aware formulation, whereas HSMSR-IVA represents the opposite extreme of an identical structural decomposition across states.
    
    \subsubsection{Training and Hyperparameter Selection}
    
    All models used \(K=20\) SCVs and were optimized for 500 epochs using Adam with learning rate \(10^{-4}\). Demixing matrices were initialized with random orthonormal rows obtained using QR decomposition, and row normalization was applied throughout optimization. For each run, the checkpoint with the lowest optimization loss was retained.
    
    The Kotz parameters were fixed at \(\lambda=1\), \(\beta=0.5\), and \(\eta=1\), with \(\epsilon_{\mathrm K}=10^{-6}\). The IP-IVA log-volume weight was \(\rho_{\mathrm V}=1\). MSR-IVA used \(\rho_g=\rho_f=1\), and \(\epsilon_{\mathrm V}=10^{-8}\) was used for log-volume stabilization. The structural residual weight was evaluated over \(\alpha\in\{0.01,0.05,0.1,0.5,1,2,5,10,25\}\), with the separately implemented no sharing model included as the unconstrained reference. We selected \(\alpha=0.1\) as a representative moderate value from the high-performing positive-\(\alpha\) regime and repeated the main comparison using seeds \(42,43,44,45,\) and \(46\).
    
    \begin{figure*}[t]
    \centering
    \includegraphics[width=0.90\textwidth]{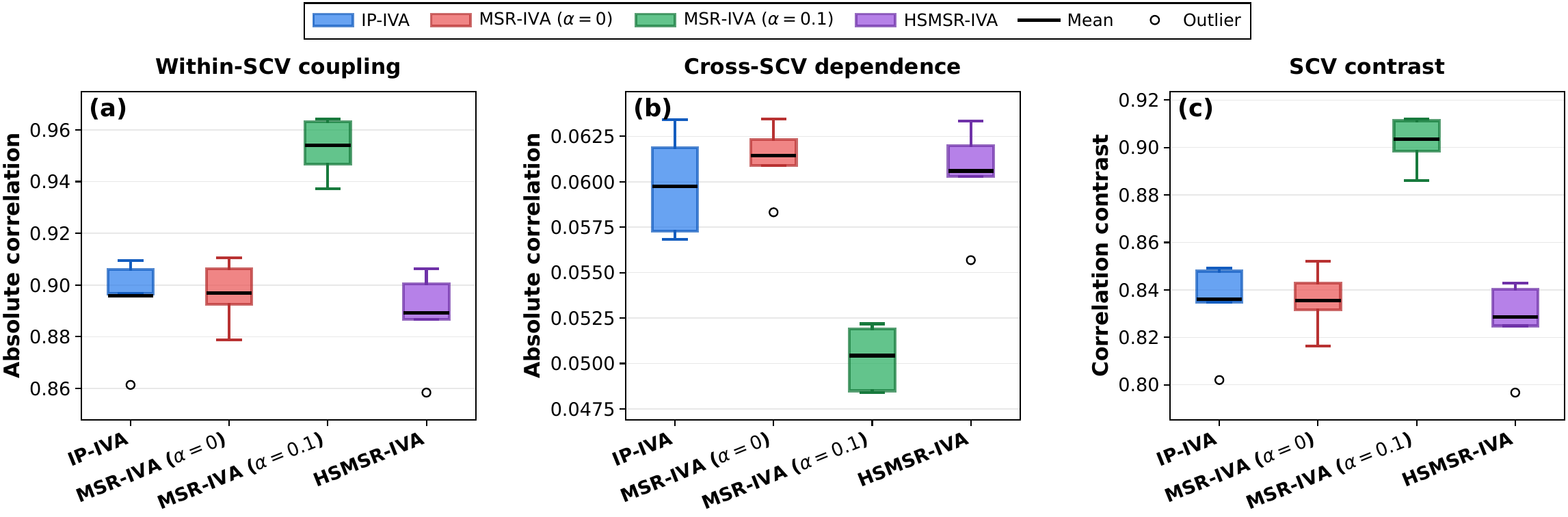}
    
    \vspace{0.07in}
    
    \includegraphics[width=0.90\textwidth]{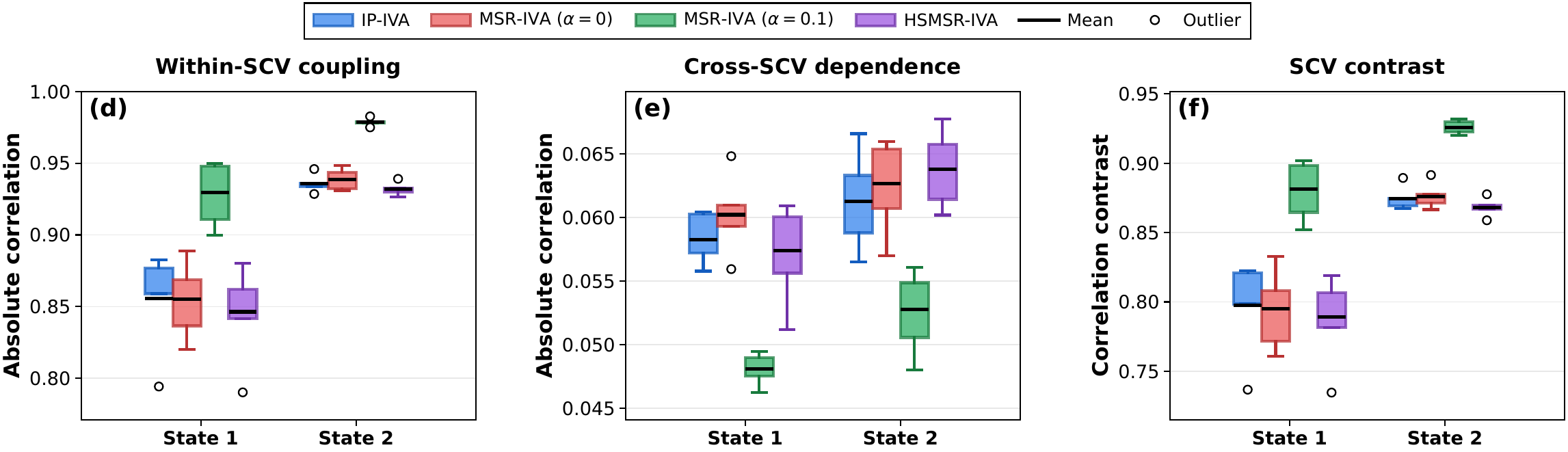}
    
    \caption{Repeated-run performance across five initialization seeds. (a--c) Performance averaged across the two dynamic states for within-SCV coupling, cross-SCV dependence, and coupling contrast. (d--f) Corresponding state-specific results for State~1 and State~2.}
    \label{fig:s2_repeated_runs}
    \end{figure*}

    
    \section{Results}

    \subsection{Sensitivity to Structural Residual Regularization}
    \label{subsec:sensitivity_ablation}
    
    Figure~\ref{fig:s2_sensitivity_ablation} shows that introducing moderate residual regularization increased matched source coupling and reduced cross-SCV dependence relative to the no sharing reference. Stronger regularization eventually reduced performance as the structural representations became increasingly constrained toward sharing. At \(\alpha=0.1\), MSR-IVA achieved \(C_{\mathrm{within}}^{(1)}=0.9389\), \(C_{\mathrm{within}}^{(2)}=0.9780\), \(C_{\mathrm{cross}}^{(1)}=0.0490\), and \(C_{\mathrm{cross}}^{(2)}=0.0480\). The corresponding coupling contrasts were \(0.8899\) for State~1 and \(0.9300\) for State~2. These results supported using \(\alpha=0.1\) as a representative value for the repeated-run comparison.

    \subsection{Robustness Across Initializations}
    \label{subsec:repeated_runs}
    
    Using \(\alpha=0.1\), MSR-IVA was compared with IP-IVA, no sharing, and hard sharing across five random initializations. Figure~\ref{fig:s2_repeated_runs} summarizes the results. Averaged across states, MSR-IVA achieved \(C_{\mathrm{within}}=0.9540\pm0.0116\), \(C_{\mathrm{cross}}=0.0504\pm0.0019\), and \(C_{\mathrm{contrast}}=0.9035\pm0.0113\). IP-IVA obtained \(0.8957\pm0.0198\), \(0.0598\pm0.0029\), and \(0.8360\pm0.0198\), respectively. The no sharing and hard sharing variants produced lower coupling contrasts of \(0.8354\pm0.0134\) and \(0.8286\pm0.0191\). MSR-IVA achieved state-specific coupling contrasts of \(0.8813\pm0.0219\) for State~1 and \(0.9258\pm0.0051\) for State~2. Relative to IP-IVA, MSR-IVA increased mean within-SCV coupling by 6.5\% and reduced mean cross-SCV dependence by 15.7\%.

    \subsection{State-Conditioned Structural Adaptation}
    \label{subsec:structural_adaptation}
    
    We further tested whether the soft sharing formulation actually produced related but nonidentical structural representations. To avoid differences in the subjects used for this comparison, the analysis was restricted to the 124 subjects expressing both dynamic states. For each method and initialization, State~1 and State~2 structural sources were optimally matched using absolute cross-state correlation. As shown in Table~\ref{tab:structural_adaptation}, independently estimated no sharing decompositions produced cross-state similarity of \(0.2978\pm0.0075\), whereas hard sharing produced identical structural sources with similarity of \(1.0000\). MSR-IVA occupied an intermediate regime at \(0.9177\pm0.0081\), showing that the residual formulation can regulate structural sharing while maintaining source correspondence and allowing deviations from identity.
    
    Importantly, all 20 MSR-IVA structural SCVs retained their same-index correspondence under optimal matching for all five random initializations. The observed adaptation therefore did not arise from arbitrary source permutation: corresponding structural SCVs remained aligned across the two states while their subject-level representations were nonidentical.
    
    \begin{table}[t]
    \centering
    \caption{Cross-state structural source similarity on the 124 common subjects (mean \(\pm\) SD across five seeds).}
    \label{tab:structural_adaptation}
    \scriptsize
    \renewcommand{\arraystretch}{1.15}
    \setlength{\tabcolsep}{7pt}
    
    \begin{tabular}{|c|c|}
    \hline
    \textbf{Method} & \textbf{Structural similarity} \\
    \hline
    No sharing   & \(0.2978 \pm 0.0075\) \\
    \hline
    MSR-IVA      & \(0.9177 \pm 0.0081\) \\
    \hline
    Hard sharing & \(1.0000 \pm 0.0000\) \\
    \hline
    \end{tabular}
    
    \end{table}
    
    \subsection{Source Coupling Analysis}
    \label{subsec:source_coupling_analysis}
    
    We further examined source correspondence using MSR-IVA with \(\alpha=0.1\) and the IP-IVA baseline. Table~\ref{tab:source_coupling_results} summarizes the state-specific coupling metrics. Across states, MSR-IVA increased mean within-SCV coupling from \(0.8966\) to \(0.9584\), reduced mean cross-SCV dependence from \(0.0619\) to \(0.0485\), and increased mean coupling contrast from \(0.8348\) to \(0.9099\).
    
    \begin{table}[t]
    \centering
    \caption{Source coupling metrics for the seed-42 comparison.}
    \label{tab:source_coupling_results}
    \scriptsize
    \renewcommand{\arraystretch}{1.15}
    \setlength{\tabcolsep}{3.5pt}
    
    \begin{tabular}{|c|c|c|c|c|}
    \hline
    \textbf{State} & \textbf{Method} &
    \(\mathbf{C_{\mathrm{within}}\uparrow}\) &
    \(\mathbf{C_{\mathrm{cross}}\downarrow}\) &
    \(\mathbf{C_{\mathrm{contrast}}\uparrow}\) \\
    \hline
    State 1 & IP-IVA  & 0.8589 & 0.0604 & 0.7985 \\
    \hline
    State 1 & MSR-IVA & \textbf{0.9389} & \textbf{0.0490} & \textbf{0.8899} \\
    \hline
    State 2 & IP-IVA  & 0.9343 & 0.0633 & 0.8710 \\
    \hline
    State 2 & MSR-IVA & \textbf{0.9780} & \textbf{0.0480} & \textbf{0.9300} \\
    \hline
    \end{tabular}
    
    \end{table}

    SCV label permutation analysis further showed that the strong diagonal coupling reflected specific source correspondence rather than arbitrary SCV pairing. The observed mean diagonal correlation substantially exceeded the mismatched SCV null distribution in State~1 (\(0.9389\) vs.\ \(0.0491\pm0.0083\)) and State~2 (\(0.9780\) vs.\ \(0.0482\pm0.0079\)). With \(10{,}000\) permutations and the plus-one correction, the global empirical value was \(p=1/10001\approx1.0\times10^{-4}\) for both states. All 20 SCVs in each state also attained the minimum max-statistic family-wise error corrected value, \(p_{\mathrm{FWER}}=1/10001\).

    
    \section{Discussion}
    \label{sec:discussion}
    
    MSR-IVA achieved the intended soft sharing behavior between independent and hard-shared structural decompositions. Cross-state structural similarity was \(0.9177\), compared with \(0.2978\) for no sharing and \(1.0000\) for hard sharing. Rather than indicating that higher similarity is inherently better, this intermediate value shows that the residual formulation can maintain cross-state correspondence without forcing identical structural representations. MSR-IVA also improved matched source coupling and reduced cross-SCV dependence relative to independent pairwise IVA.
    
    Because structural similarity was evaluated on the same 124 subjects expressing both states, the observed differences are not explained by different evaluation cohorts. The masking formulation further allows each state to use all available observations instead of restricting fusion to the common-subject intersection. This ability to preserve matched structural--functional relationships while retaining state-specific variation may be useful for downstream tasks such as biomarker identification and classification, which will be investigated in future work.

    
    \section{Conclusion}
    \label{sec:conclusion}
    
    We proposed MSR-IVA, a state-aware framework for fusing one structural modality with multiple partially observed dFNC states. By combining a shared structural component with state-specific residuals and state masks, MSR-IVA enables controllable structural sharing while retaining available observations for each state.
    
    Across five random initializations, MSR-IVA improved matched source coupling by 6.5\% and reduced unmatched source dependence by 15.7\% relative to IP-IVA. The structural similarity results further demonstrated an intermediate sharing regime between independent and identical decompositions, supporting the intended behavior of the proposed formulation.
    
    
    \section{Acknowledgment}
    
    This work is supported by the National Institutes of Health under grant NIH R01AG090597. The authors gratefully acknowledge this support. The MSR-IVA implementation is publicly available at \href{https://github.com/dasjar/MSRIVA}{https://github.com/dasjar/MSRIVA}.

    
    \bibliographystyle{IEEEbib}
    \bibliography{refs}

@article{adali2015multimodal,
  title={Multimodal data fusion using source separation: Two effective models based on ICA and IVA and their properties},
  author={Adali, T{\"u}lay and Levin-Schwartz, Yuri and Calhoun, Vince D.},
  journal={Proceedings of the IEEE},
  volume={103},
  number={9},
  pages={1478--1493},
  year={2015},
  doi={10.1109/JPROC.2015.2461601}
}

@inproceedings{kim2006iva,
  title={Independent Vector Analysis: An Extension of ICA to Multivariate Components},
  author={Kim, Taesu and Eltoft, Torbj{\o}rn and Lee, Te-Won},
  booktitle={Independent Component Analysis and Blind Signal Separation},
  series={Lecture Notes in Computer Science},
  volume={3889},
  pages={165--172},
  publisher={Springer},
  year={2006},
  doi={10.1007/11679363_21}
}

@article{allen2014tracking,
  title={Tracking Whole-Brain Connectivity Dynamics in the Resting State},
  author={Allen, Elena A. and Damaraju, Eswar and Plis, Sergey M. and Erhardt, Erik B. and Eichele, Tom and Calhoun, Vince D.},
  journal={Cerebral Cortex},
  volume={24},
  number={3},
  pages={663--676},
  year={2014},
  doi={10.1093/cercor/bhs352}
}

@article{silva2021misa,
  title={Multidataset Independent Subspace Analysis With Application to Multimodal Fusion},
  author={Silva, Rogers F. and Plis, Sergey M. and Adali, T{\"u}lay and Pattichis, Marios S. and Calhoun, Vince D.},
  journal={IEEE Transactions on Image Processing},
  volume={30},
  pages={588--602},
  year={2021},
  doi={10.1109/TIP.2020.3028452}
}

@article{jack2008alzheimer,
  title={The Alzheimer's disease neuroimaging initiative (ADNI): MRI methods},
  author={Jack Jr, Clifford R and Bernstein, Matt A and Fox, Nick C and Thompson, Paul and Alexander, Gene and Harvey, Danielle and Borowski, Bret and Britson, Paula J and L. Whitwell, Jennifer and Ward, Chadwick and others},
  journal={Journal of Magnetic Resonance Imaging: An Official Journal of the International Society for Magnetic Resonance in Medicine},
  volume={27},
  number={4},
  pages={685--691},
  year={2008},
  publisher={Wiley Online Library}
}

@article{du2020neuromark,
  title={NeuroMark: An automated and adaptive ICA based pipeline to identify reproducible fMRI markers of brain disorders},
  author={Du, Yuhui and Fu, Zening and Sui, Jing and Gao, Shuang and Xing, Ying and Lin, Dongdong and Salman, Mustafa and Abrol, Anees and Rahaman, Md Abdur and Chen, Jiayu and others},
  journal={NeuroImage: Clinical},
  volume={28},
  pages={102375},
  year={2020},
  publisher={Elsevier}
}

@article{du2018neuromark,
  title={NeuroMark: An automated and adaptive ICA based pipeline to identify reproducible fMRI markers of brain disorders},
  author={Du, Yuhui and Fu, Zening and Sui, Jie and others},
  journal={NeuroImage},
  volume={172},
  pages={566--584},
  year={2018},
  publisher={Elsevier}
}

@article{sendi2021alzheimer,
  title={Alzheimer’s disease projection from normal to mild dementia reflected in functional network connectivity: a longitudinal study},
  author={Sendi, Mohammad SE and Zendehrouh, Elaheh and Miller, Robyn L and Fu, Zening and Du, Yuhui and Liu, Jingyu and Mormino, Elizabeth C and Salat, David H and Calhoun, Vince D},
  journal={Frontiers in Neural Circuits},
  volume={14},
  pages={593263},
  year={2021},
  publisher={Frontiers Media SA}
}

@article{solomon2025timeseries,
  author  = {Solomon, Victor and Wen, Junzhi and Angryk, Rafal and Rampuria, Manya and Rayala, Omkar and Afrid, Abdul},
  title   = {Time Series Decomposition Using Wavelet and {Fourier} Transforms for Enhanced Solar Flare Forecasting},
  journal = {The International FLAIRS Conference Proceedings},
  volume  = {38},
  number  = {1},
  year    = {2025},
  month   = may,
  doi     = {10.32473/flairs.38.1.139016},
  url     = {https://journals.flvc.org/FLAIRS/article/view/139016}
}
    
    \end{document}